\documentclass{article} 
\usepackage{iclr2021_conference,times}

\usepackage{amsmath,amsfonts,bm}

\def\eqref#1{equation~\ref{#1}}

\def\1{\bm{1}}

\DeclareMathAlphabet{\mathsfit}{\encodingdefault}{\sfdefault}{m}{sl}
\SetMathAlphabet{\mathsfit}{bold}{\encodingdefault}{\sfdefault}{bx}{n}

\usepackage{hyperref}
\usepackage{url}

\title{Compositional Generalization with Tree Stack Memory Units}

\usepackage[utf8]{inputenc} 
\usepackage[T1]{fontenc}    
\usepackage{hyperref}       
\usepackage{url}            
\usepackage{booktabs}       
\usepackage{amsfonts}       
\usepackage{nicefrac}       
\usepackage{microtype}      
\usepackage{wrapfig}

\usepackage{epsf,psfrag,amssymb,amsfonts,latexsym,graphicx,bm,xcolor,array}
\usepackage{subcaption}
\usepackage{float}
\usepackage{verbatim}
\usepackage[mathscr]{eucal}
\usepackage[tbtags]{mathtools}
\usepackage{multirow}
\usepackage{amsmath}
\usepackage{enumitem}
\usepackage{thmtools,thm-restate}
\usepackage{algorithm}
\usepackage{algorithmic}
\usepackage{listings}
\usepackage{wrapfig}
\usepackage{tabularx}
\usepackage{multicol}
\usepackage{makecell}

\lstdefinestyle{customPython}{
  belowcaptionskip=1\baselineskip,
  breaklines=true,
  frame=L,
  xleftmargin=\parindent,
  language=Python,
  showstringspaces=false,
  basicstyle=\footnotesize\ttfamily\scriptsize,
  keywordstyle=\bfseries\color{green!40!black},
  commentstyle=\sffamily\fontseries{lc}\itshape\color{purple!40!black},
  identifierstyle=\color{blue},
  stringstyle=\color{orange},
}

\DeclareMathOperator{\push}{push}
\DeclareMathOperator{\pop}{pop}

\DeclareMathOperator{\noop}{no-op}
\DeclareMathOperator{\actionPush}{\mathbf{a}_{j}^{\push}}
\DeclareMathOperator{\actionPop}{\mathbf{a}_{j}^{\pop}}
\DeclareMathOperator{\actionNoop}{\mathbf{a}_{j}^{\noop}}
 
\definecolor{brickred}{rgb}{0.8, 0.25, 0.33}

\newcommand{\cut}[1]{}

\newcommand{\SMU}{SMU}
\newcommand{\SMUlong}{Tree Stack Memory Unit}

\author{%
 Forough Arabshahi\thanks{Equal Contribution} \\
 Carnegie Mellon University\\
 Pittsburgh, PA 15213 \\
 \texttt{farabsha@cs.cmu.edu} \\
  \And
  Zhichu Lu$^*$ \\
 Johns Hopkins University\\
 Baltimore, MD 21218 \\
 \texttt{zhicul@cs.cmu.edu} \\
  \And
 Pranay Mundra \\
 University of Washington \\
 Seattle, WA  \\
 \texttt{pranay99@uw.edu} \\
  \And
 Sameer Singh \\
 University of California Irvine \\
 Irvine, CA 92697 \\
 \texttt{sameer@uci.edu} \\
 \And
 Animashree Anandkumar \\
 California Institute of Technology \\
 Pasadena, CA 91125 \\
 \texttt{anima@caltech.edu} \\
}

\iclrfinalcopy 
\begin{document}

\maketitle

\begin{abstract}
    We study compositional generalization, viz., the problem of zero-shot generalization to novel compositions of   concepts in a domain. Standard neural networks fail to a large extent on  compositional learning.
    We propose \SMUlong s (Tree-\SMU) to enable strong compositional generalization. 
    Tree-\SMU~is a recursive neural network with Stack Memory Units (\SMU s), a novel memory augmented neural network whose memory has a differentiable stack structure. Each \SMU~in the tree architecture learns to read from its stack and to write to it by combining the stacks and states of its children   through gating. 
    The stack helps capture long-range  dependencies in the problem domain, thereby enabling compositional generalization. Additionally, the  stack also preserves the ordering of each node's descendants, thereby retaining \emph{locality} on the tree.
    We demonstrate strong empirical results on two mathematical reasoning benchmarks. We use four   compositionality tests to assess the generalization performance of Tree-\SMU~and show that it enables accurate compositional generalization compared to strong baselines such as Transformers and Tree-LSTMs.
\end{abstract}

\vspace{-2em}
\section{Introduction}

Despite the impressive performance of deep learning in the last decade, systematic compositional generalization is mostly  out of reach for standard neural networks~\citep{hupkes2020compositionality}. 
In a compositional domain, a set of concepts can be combined in novel ways to form new instances. Compositional generalization is defined as zero-shot generalization to such novel compositions.~\citet{lake2018generalization} recently  showed that a variety of recurrent neural networks fail spectacularly on tasks requiring compositional generalization. ~\citet{lample2019deep} presented similar results for the popular transformer architecture and  \citet{vaswani2017attention} showed it in the symbolic mathematical domain. In particular, they showed that transformers can generalize nearly perfectly when the training and test distributions are identical. However, this generalization is brittle and  their performance degrades significantly even with slight distributional shifts.


Neuro-symbolic models, on the other hand, hold the promise for achieving compositional generalization~\citep{lamb2020graph}. They   integrate symbolic domain knowledge 
within neural architectures. For instance, a popular example is a recursive (tree-structured) neural network with nodes corresponding to different symbolic concepts and the tree representing their composition. Neuro-symbolic models can achieve better generalization since the neural component is relieved from the additional burden of learning symbolic knowledge from scratch. Moreover, in many domains, compositional supervision is readily available, such as the domain of mathematical equations, and neuro-symbolic models can directly incorporate it~\citep{allamanis2017learning,arabshahi2018combining,evans2018can}.   



Recursive neural networks, however, still fall short when it comes to compositional generalization, especially on instances much more complex (larger tree depth) compared to  training data. This is because of error propagation along the tree, and the failure of recursive networks to capture long-range dependencies effectively. There are currently no error-correction mechanisms in recursive neural networks to overcome this. We address this issue in this paper by designing a novel recursive neural network (Tree-\SMU) with stack memory units.

In order to evaluate the compositional generalization of different neural network models, we use the  tests proposed by~\citet{hupkes2020compositionality}. The tests evaluate the model on its ability to (1) generalize systematically to novel compositions, (2) generalize to deeper compositions than they are trained on, (3) generalize to shallower compositions than they are trained on, and (4) learn similar representations for semantically equivalent compositions.

Mathematical reasoning is an excellent test bed for such evaluations of compositional generalization, since we can construct arbitrarily deep or shallow compositions using a given set of primitive functions in mathematics. Moreover, there has recently been a growing interest in the problem of mathematical reasoning~\citep{lee2019mathematical,saxton2019analysing,lample2019deep,arabshahi2018combining,loos2017deep,allamanis2017learning}. 
Furthermore, it is difficult (and often ambiguous) to accurately measure compositionality in other domains such as natural language or vision; benchmarks that do measure compositionality in these domains are synthetic datasets, which are far from free-form natural language or real images~\citep{lake2018generalization,johnson2017clevr,hupkes2020compositionality}.

\begin{figure*}
    \centering
    \begin{subfigure}[t]{0.23\textwidth}{
    \includegraphics[width=\textwidth]{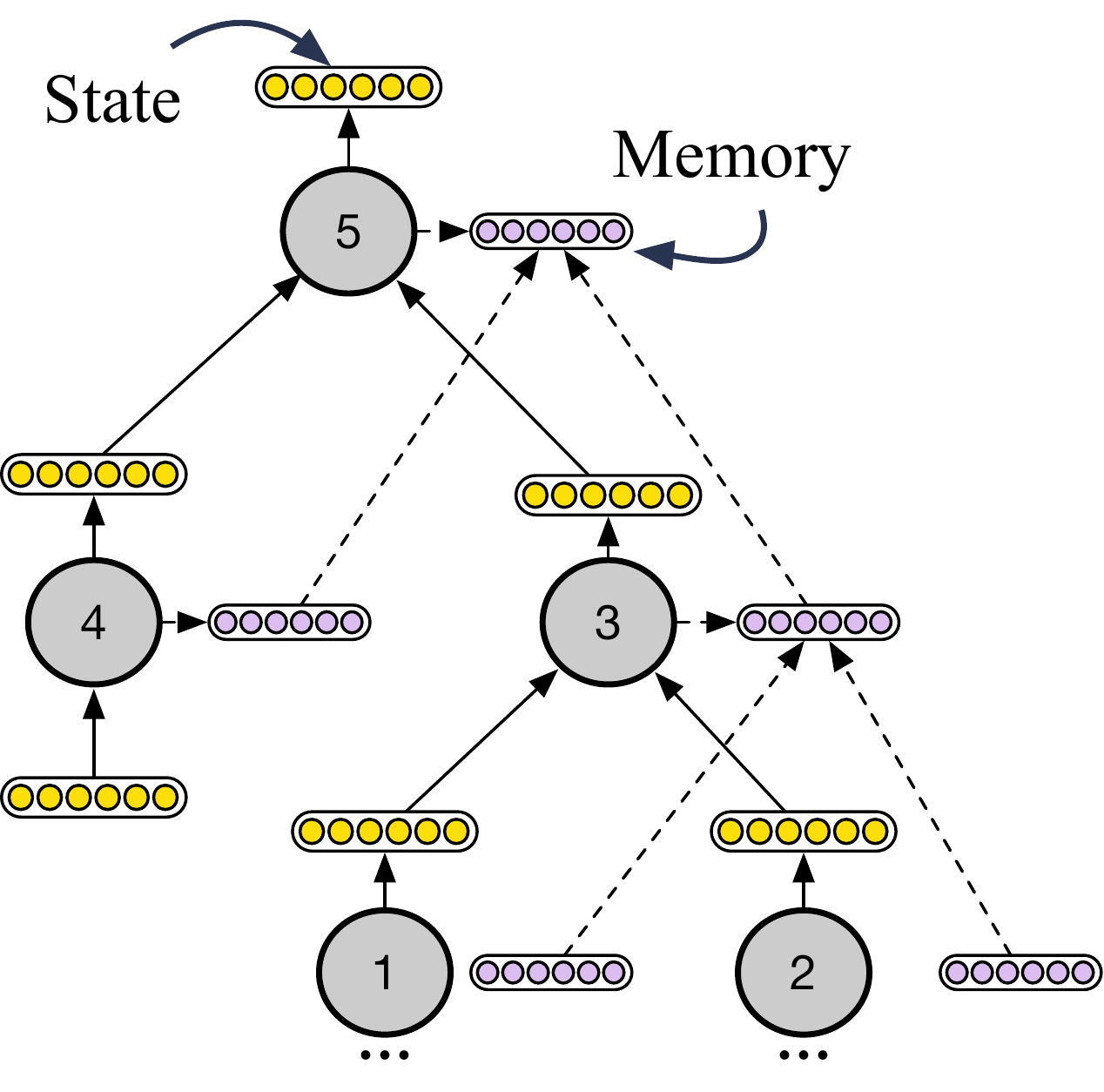}}
    \caption{Tree-LSTM}
    \label{fig:model_TLSTM}
    \end{subfigure}
    ~
    \begin{subfigure}[t]{0.24\textwidth}{
        \includegraphics[width=\textwidth]{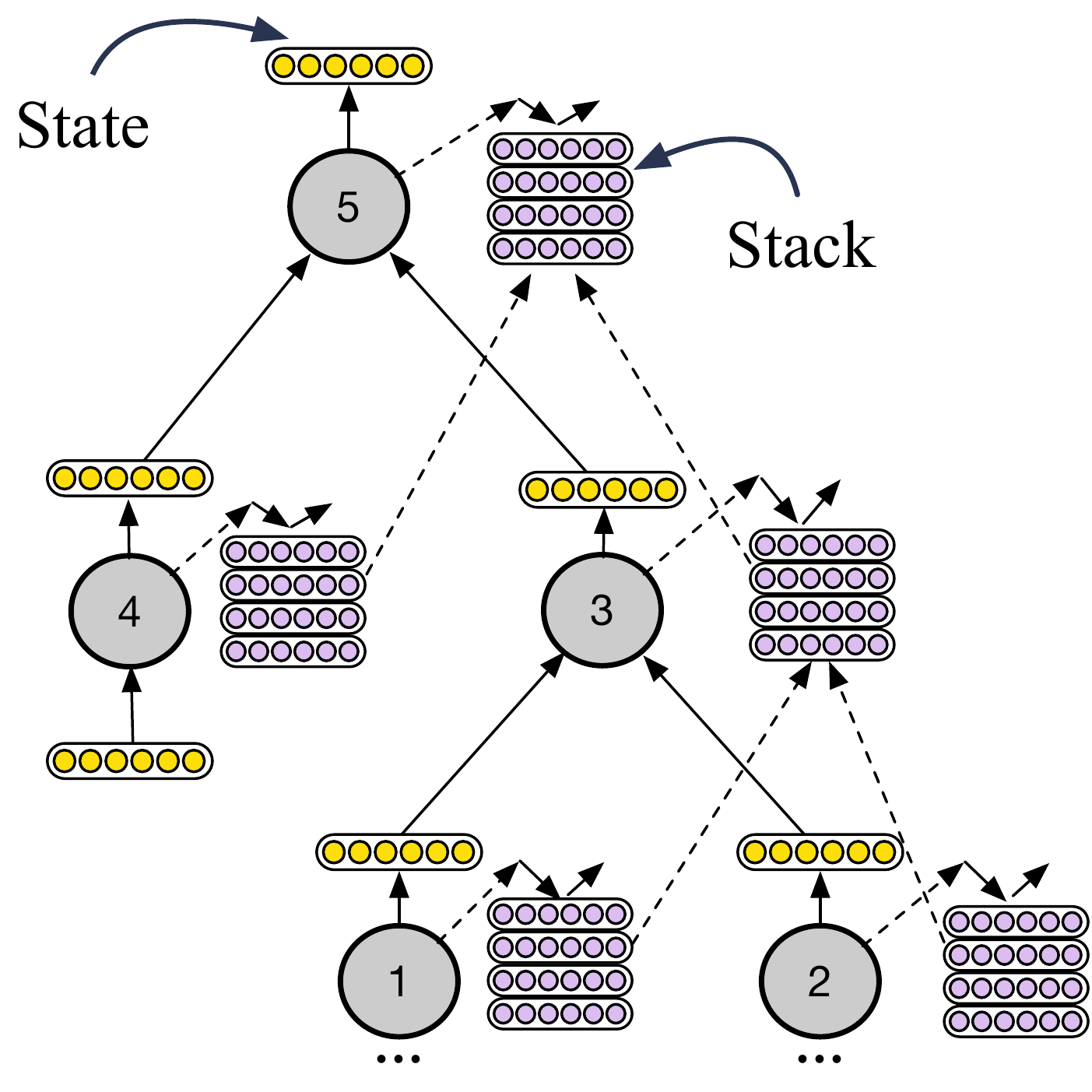}}
    \caption{Tree-\SMU}
    \label{fig:model_TLSTMstack}
    \end{subfigure}
    \begin{subfigure}[t]{0.35\textwidth}{
            \includegraphics[width=\textwidth]{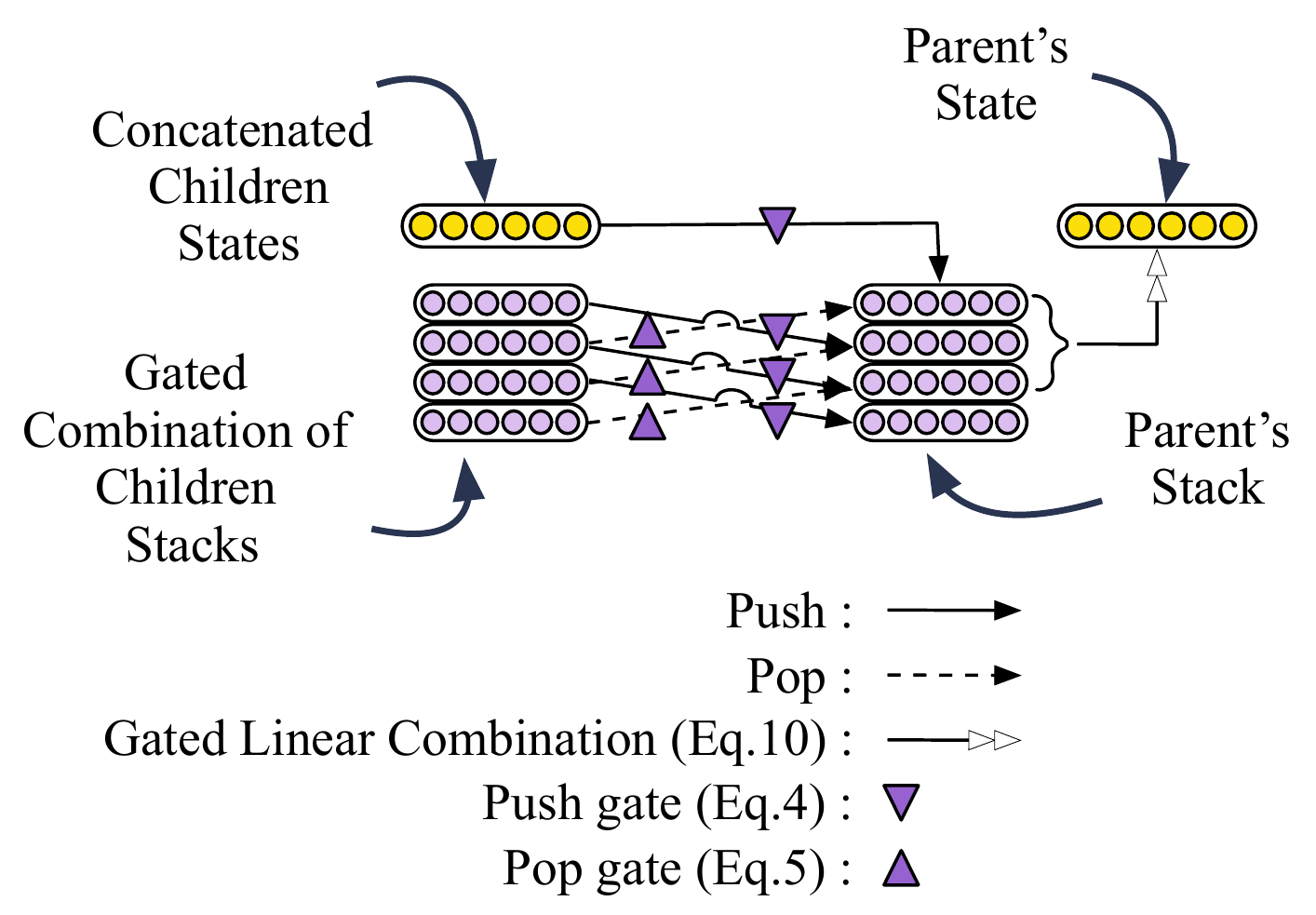}}
        \caption{Soft Push and Pop}
        \label{fig:pushpop}
    \end{subfigure}
    \caption{Model architecture of Tree-LSTM vs. Tree-\SMU~(Tree Stack Memory Unit). Compared to LSTM, a SMU has an increased memory capacity of the form of a 
    differentiable stack. The black arrows over the stacks in Fig.~\ref{fig:model_TLSTMstack} represent our soft push/pop operation, shown in Fig.~\ref{fig:pushpop}. As depicted in Fig.~\ref{fig:pushpop}, the children's stacks and states jointly determine the content of the parent stack. The children stacks are combined using the gating mechanism in Equation~\ref{eq:stackMem} and then used to fill the parent stack through the push and pop gating operations given in Equations \ref{eq:lstmstack0} and \ref{eq:lstmstacki}.
    }
    \label{fig:model}
    \vspace{-1.5em}
\end{figure*}

    
    
\begin{wrapfigure}{r}{0.25\textwidth}
    \begin{center}
    \includegraphics[width=0.25\textwidth]{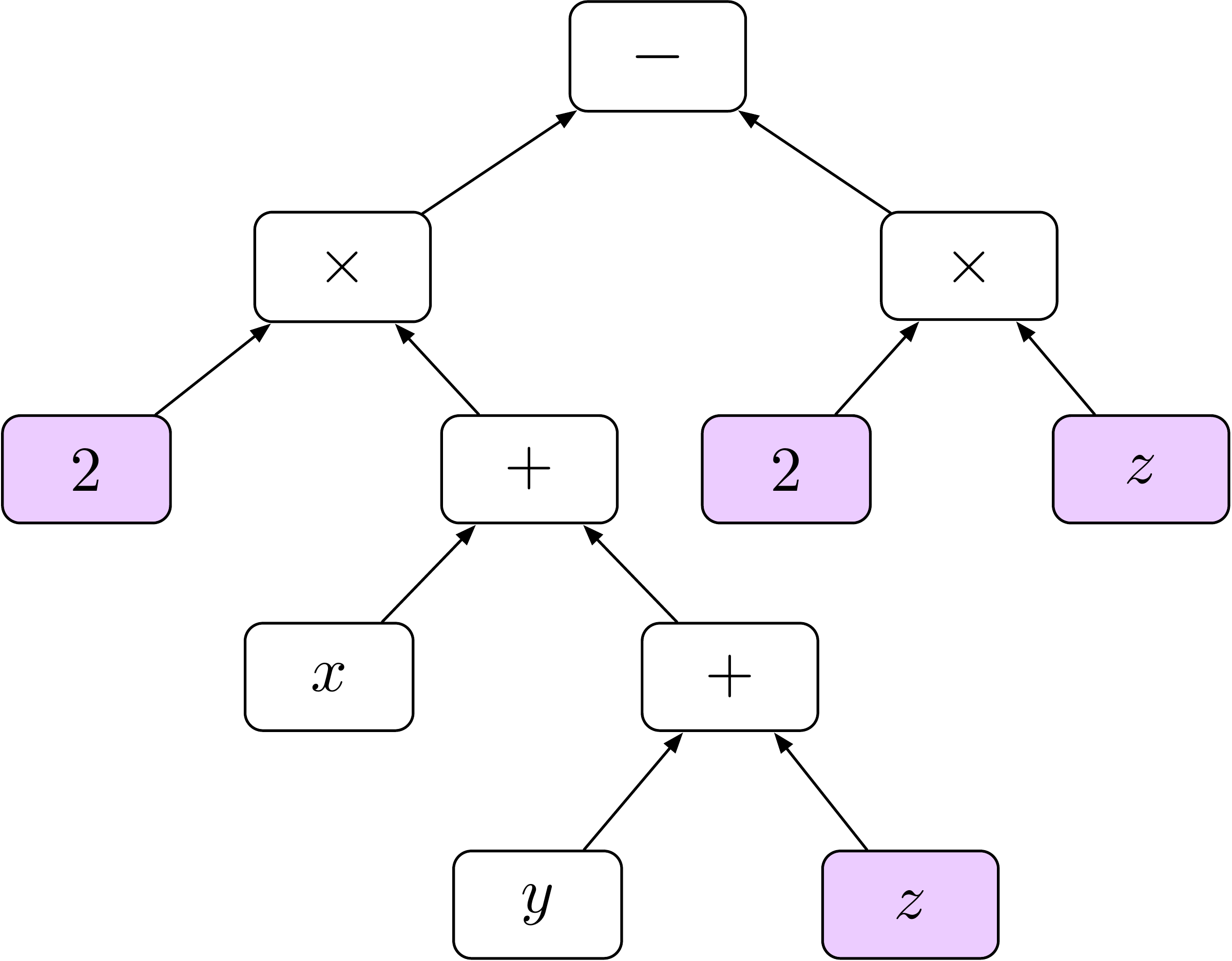}
    \end{center}
    \caption{Expression tree for $2\big(x+(y+z)\big)-2z$. Because the equation is evaluated bottom-up, having access to descendants $2$ and $z$ at the tree root enables the model to correctly evaluate the expression.}
    \label{fig:long_range_eq}
    \vspace{-2em}
\end{wrapfigure}
\vspace{-1em}
\paragraph{Summary of Results} In this paper,
\begin{itemize}[leftmargin=4ex, topsep=-1ex, itemsep=-0ex]
    \item[1] We propose a novel recursive neural network architecture, \SMUlong~(Tree-\SMU), to enable compositional generalization in the domain of mathematical reasoning.
    \item[2] We evaluate  generalization of Tree-\SMU~on four different compositionality tests. We show that Tree-\SMU~consistently outperforms the compositional generalization of powerful baselines such as transformers, tree transformers~\citep{shiv2019novel} and Tree-LSTMs.
    \item[3] We observe that improving the zero-shot generalization is also correlated with improved sample efficiency of training.
\end{itemize}

We propose augmenting the memory of recursive neural networks with a stack, thereby enabling them to capture long-range dependencies.

\textbf{Example to show how stacks can capture non-local dependency: }Consider evaluating the mathematical expression $2\big(x+(y+z)\big) - 2z$ in Fig \ref{fig:long_range_eq}. 
At the subtract node (root), the model must have access to the representations of $2$ and $z$ to be able to correctly equate the expression to $2x + 2y$. 
However, the nodes of a Tree-RNN or Tree-LSTM only have access to the states of their direct children, as imposed by the tree structure. Thus the representations of $2$ and $z$ will be mixed together by the time they reach the subtract node. Since the models do not learn perfect representations, the mixed representation of $2$ and $z$ is inaccurate and the models have difficulty correctly evaluating the expression.
On the other hand, a neural model with an extended memory capacity, such as a Tree-\SMU, can store intermediate representations at each node. If the memory is correctly used and trained, the subtract node in Fig~\ref{fig:long_range_eq}, will have access to the original, unmixed representations of $2$ and $z$ and can correctly evaluate the equation.

We propose \SMUlong~(Tree-\SMU), a tree-structured neural network whose nodes (\SMU s) have an extended memory capacity with the structure of a differentiable stack (Fig.~\ref{fig:model}). The stack memory is built into the architecture of each \SMU~node, which learns to read (pop) from and write (push) to it. The nodes push the gated combination of their children's states and stacks onto their own stacks (Fig.~\ref{fig:pushpop}). Since, the intermediate results of the descendants are stored in the children's stacks, the current node also gets access to its descendants' states if it chooses to push. This error-correction mechanism allows the tree model to capture long-range dependencies. 

We choose the stack data structure (over say, a queue) because it preserves the ordering of each node's descendants. Therefore, this design retains \emph{locality} while allowing for learning of \emph{global} representations. It is worth noting that although Tree-\SMU~has a larger memory capacity, the model size is itself about the same as a tree-LSTM. Moreover, we show that merely increasing the model size in a tree-LSTM does not improve compositional generalization since it leads to overfitting to the training data distribution. Thus, we need sophisticated error-correction mechanism of Tree-\SMU~to capture long-range dependency effectively for compositional generalization.

We test our model on two mathematical reasoning tasks, viz., equation verification and equation completion~\citep{arabshahi2018combining}. On equation completion, we show through t-SNE visualization that Tree-\SMU~learns a much smoother representation for mathematical expressions, whereas Tree-LSTM is more sensitive to irrelevant syntactic details. On equation completion, Tree-\SMU~achieves $6\%$ better top-5 accuracy compared to Tree-LSTM. On the equation verification task, Tree-\SMU~achieves 7\% accuracy improvement for generalizing to shallow equations and 2\% accuracy improvement for generalizing to deeper equations, compared to Tree-LSTM, and 18.5\% compared to transformers and 17.5\% compared to tree transformers. For generalization at the same tree depth, we obtain 6.8\% accuracy improvement compared to transformer and 5.5\% compared to tree transformers with slightly (<1\%) better accuracy compared to Tree-LSTM. Thus, we demonstrate that Tree-\SMU~ achieves the state-of-art performance, across the board on different compositionality tests. 

\vspace{-1em}
\section{Background and Notation}
\label{sec:background}
\vspace{-1em}
A recursive neural network is a tree-structured network in which
each node is a neural network block. The tree's structure is often imposed by the data. For example, in mathematical equation verification the structure is that of the input equation (e.g., Figure \ref{fig:long_range_eq}). For simplicity, we present the formulation of binary recursive neural networks. However, all the formulations can be trivially extended to n-ary recursive neural networks and child-sum recursive neural networks. 
We present matrices with bold uppercase letters, vectors with bold lowercase letters, and scalars with non-bold letters. 

All the nodes (blocks) of a recursive neural network have a state, $\mathbf{h}_j \in \mathbb{R}^n$, and an input, $\mathbf{i}_j \in \mathbb{R}^{2n}$, where $n$ is the hidden dimension, $j \in [0,N-1]$ and $N$ is the number of nodes in the tree. We label the children of node $j$ with $c_{j1}$ and $c_{j2}$. We have $\mathbf{i}_j = [\mathbf{h}_{c_{j1}};\mathbf{h}_{c_{j2}}] \label{eq:catH}$
where $[\cdot \; ; \, \cdot]$ indicates concatenation. If the block is a leaf node, $\mathbf{i}_j$ is the external input of the network. For example, in the equation tree shown in Figure \ref{fig:long_range_eq}, all the terminal nodes (leaves) are the external neural network inputs (represented with 1-hot vectors, for example). For simplicity we assume that non-terminal nodes do not have an external input and only take inputs from their children. However, if they exist, we can easily handle them by
additionally concatenating them with the children's states at each node's input. 
The way $\mathbf{h}_j$ is computed using $\mathbf{i}_j$ depends on the neural network block's architecture. For example, in a Tree-RNN and Tree-LSTM, $\mathbf{h}_j$ is computed by passing $\mathbf{i}_j$ through a feedforward neural network and a LSTM network, respectively. Each node in Tree-LSTM also has a 1-D memory vector (Fig.~\ref{fig:model_TLSTM})

\vspace{-1em}
\section{\SMUlong~(Tree-\SMU)}
\label{sec:method}
\vspace{-1em}
In this section, we introduce \SMUlong s~(Tree-\SMU) that consist of SMU nodes which incorporate a differentiable stack as a memory (Fig.~\ref{fig:model_TLSTMstack}). Therefore, this structure has an increased memory capacity compared to a Tree-LSTM. Each \SMU~learns to read from and write to its stack and Tree-\SMU~learns to propagate information up the tree and to fill the stacks of each node using its children's states and stacks (Fig.~\ref{fig:pushpop}). 
The states of the descendants are stored in the children's stacks. Therefore, globally, the stack enables each node of Tree-\SMU~to have indirect access to the state of its descendants. This is an error correction mechanism that captures long-range dependencies. 
Locally, stack preserves the order of each node's descendants. If a queue was used instead, the descendants' states would be stored in the reverse order. Therefore, all the nodes see the state of the tree leaves on top of their memory. This destroys locality and hurts compositional generalization. 
Despite its increased memory capacity, the number of parameters of Tree-\SMU~is similar to Tree-LSTM for the same hidden state dimension. The stack memory is filled up and emptied using our proposed soft push and pop gates. Finally, a gated combination of the children's stacks along with the concatenated states of the children determine the content of the parent stack. 
Each \SMU~node $j$ has a stack $\mathbf{S}_j \in \mathbb{R}^{p\times n}$ where $p$ is the stack size.
A stack is a LIFO data structure and the network only interacts with it through its top (in a soft way). 
We use the notation $\mathbf{S}_j[i] \in \mathbb{R}^n$ to refer to memory row $i$ for $i\in0,\dots p-1$, where $i=0$ indicates the stack top.
For each node $j$, the children's stacks are combined using the equations below 
\begin{align}
    \mathbf{S}_{c_j}[i] = \mathbf{f}_{j1} \odot \mathbf{S}_{c_{j1}}[i]& + \mathbf{f}_{j2} \odot \mathbf{S}_{c_{j2}}[i], \label{eq:stackMem} \\
    \mathbf{f}_{j1} = \sigma(\mathbf{U}_{j1}^{(f)} \mathbf{i}_j +  \mathbf{b}_{j1}^{(f)}),& \quad \mathbf{f}_{j2} = \sigma(\mathbf{U}_{j2}^{(f)} \mathbf{i}_j +  \mathbf{b}_{j2}^{(f)}). \label{eq:lstmF2} 
\end{align}
Where matrices $\mathbf{U}_{j1}^{(f)}$ and $\mathbf{U}_{j2}^{(f)}$, and vectors $\mathbf{b}_{j1}^{(f)}$ and $\mathbf{b}_{j2}^{(f)}$ are trainable weight and biases similar to the forget gates on Tree-LSTM. The push and pop gates are element-wise operators given below,
\begin{align}
    \actionPush &= \sigma(\mathbf{A}_j^{(\push)} \mathbf{i}_j + \mathbf{b}_j^{(\push)}) \label{eq:actionPush_smu},\\
    \actionPop &= \sigma(\mathbf{A}_j^{(\pop)} \mathbf{i}_j + \mathbf{b}_j^{(\pop)}) \label{eq:actionPop_smu},
\end{align}
where $\mathbf{A}_j^{(\push)}, \mathbf{A}_j^{(\pop)}, \mathbf{b}_j^{(\push)}, \mathbf{b}_j^{(\pop)}$ are trainable weights and biases and gates $\actionPush, \actionPop \in \mathbb{R}^{n}$ are element-wise normalized to 1.
The stack is initialized with 0s and its update equations are,
\begin{align}
    \mathbf{S}_j[0] =& \actionPush \odot \; \mathbf{u}_j +
    \actionPop \odot \; \mathbf{S}_{c_j}[1] \label{eq:lstmstack0},  \\
    \mathbf{S}_j[i] =& \actionPush \odot \; \mathbf{S}_{c_j}[i-1] +
    \actionPop \odot \; \mathbf{S}_{c_j}[i+1],
    \label{eq:lstmstacki}
\end{align}
where $\mathbf{u}_j$ is given below for trainable weights and biases $\mathbf{U}_j^{(u)}$ and $\mathbf{b}_j^{(u)}$,
\begin{equation}
    \mathbf{u}_j = \tanh(\mathbf{U}_j^{(u)} \mathbf{i}_j + \mathbf{b}_j^{(u)}), \label{eq:lstmU}
\end{equation}
The output state is computed by looking at the top-k stack elements as shown below if $k>1$,
\begin{align}
    \mathbf{p}_j &= \sigma(\mathbf{U}_{j}^{(p)} \mathbf{i}_j + \mathbf{b}_{j}^{(p)}), \label{eq:posGate} \\
    \mathbf{h}_j &= \mathbf{o}_j \odot \tanh\Big(\mathbf{p}_{j} \mathbf{S}_{j}[0:k-1]  \Big), \label{eq:lstmHtopK}
\end{align}
where $\mathbf{p}_j \in \mathbb{R}^{1\times k}$ and $\mathbf{U}_{j}^{(p)} \in \mathbb{R}^{k\times n},$ $ \mathbf{b}_{j}^{(p)}$ are trainable weights and biases, $\mathbf{S}_{j}[0:k-1]$ indicates the top-k rows of the stack and $k$ is a problem dependent tuning parameter. For $k=1$:
\begin{equation}
    \mathbf{h}_j = \mathbf{o}_j \odot \tanh(\mathbf{S}_{j}[0]), \label{eq:lstmHtop1}
\end{equation}
where $\mathbf{o}_j$ is given below for trainable weights and biases $\mathbf{U}_j^{(o)}$ and $\mathbf{b}_j^{(o)}$,
\begin{equation}
    \mathbf{o}_j = \sigma(\mathbf{U}_j^{(o)} \mathbf{i}_j + \mathbf{b}_j^{(o)}).
\end{equation}
\paragraph{Additional stack operation: No-Op}
We can additionally add another stack operation called no-op. No-op is the state where the network neither pushes to the stack nor pops from it and keeps the stack in its previous state.
The no-op gate and the stack update equations are given below $\actionNoop \in \mathbb{R}^{n}$. 
\begin{align}
    \actionNoop =& \sigma(\mathbf{A}_j^{(\noop)} \mathbf{i}_j + \mathbf{b}_j^{(\noop)}), \label{eq:actionPush}\\
    \mathbf{S}_j[0] =& \actionPush \odot \; \mathbf{u}_j +
    \actionPop \odot \; \mathbf{S}_{c_j}[1] +
    \actionNoop \odot \; \mathbf{S}_{c_j}[0], \label{eq:lstmstackNoOp0} \\ 
    \mathbf{S}_j[i] =& \actionPush \odot \; \mathbf{S}_{c_j}[i-1] + 
    \actionPop \odot \; \mathbf{S}_{c_j}[i+1] + \actionNoop \odot \; \mathbf{S}_{c_j}[i].
    \label{eq:lstmstackNoOpi}
\end{align}

\vspace{-1em}
\section{Experimental Setup}
\label{sec:exp}
\vspace{-1em}
In this section, we present our studied benchmark tasks, our compositional generalization tests, and experiment details.
We briefly define these tasks below. 
\begin{itemize}[leftmargin=*, topsep=0pt, itemsep=0ex]
\item[] \textbf{Mathematical Equation Verification} 
In this task, the inputs are symbolic and numeric mathematical equations from trigonometry and linear algebra and the goal is to verify their correctness. For example, the symbolic equation $1+\tan^2(\theta) = \frac{1}{\cos^2(\theta)}$ is correct
, whereas the numeric equation $\sin(\frac{\pi}{2}) = 0.5$ is incorrect. This domain contains 29 trigonometric and algebraic functions.
\item[] \textbf{Mathematical Equation Completion} In this task, the input is a mathematical equation that has a blank in it. The goal is to find a value for the blank such that the mathematical equation holds. For example for the equation $\sin^2{\theta} + \_ = 1$, the blank is $\cos^2{\theta}$.
\vspace{-0.5em}
\end{itemize}
We used the code released by \citet{arabshahi2018combining} to generate more data of significantly higher depth. We generate a dataset for training and validation purposes and another dataset with a different seed and generation hyperparameter (to encourage it to generate deeper expressions) for testing. All the equation verification models are run only once on this test set.
The train and validation data have 40k equations combined and the distribution of their depth is shown in Fig.~\ref{fig:data_stats} in the Appendix. The test dataset has around 200k equations of depth 1-19 with a pretty balanced depth distribution of about 10k equations per depth (apart from shallow expression of depth 1 and 2).
\begin{figure}
    \centering
    \begin{subfigure}{0.4\textwidth}
    \centering
    \includegraphics[width=\textwidth]{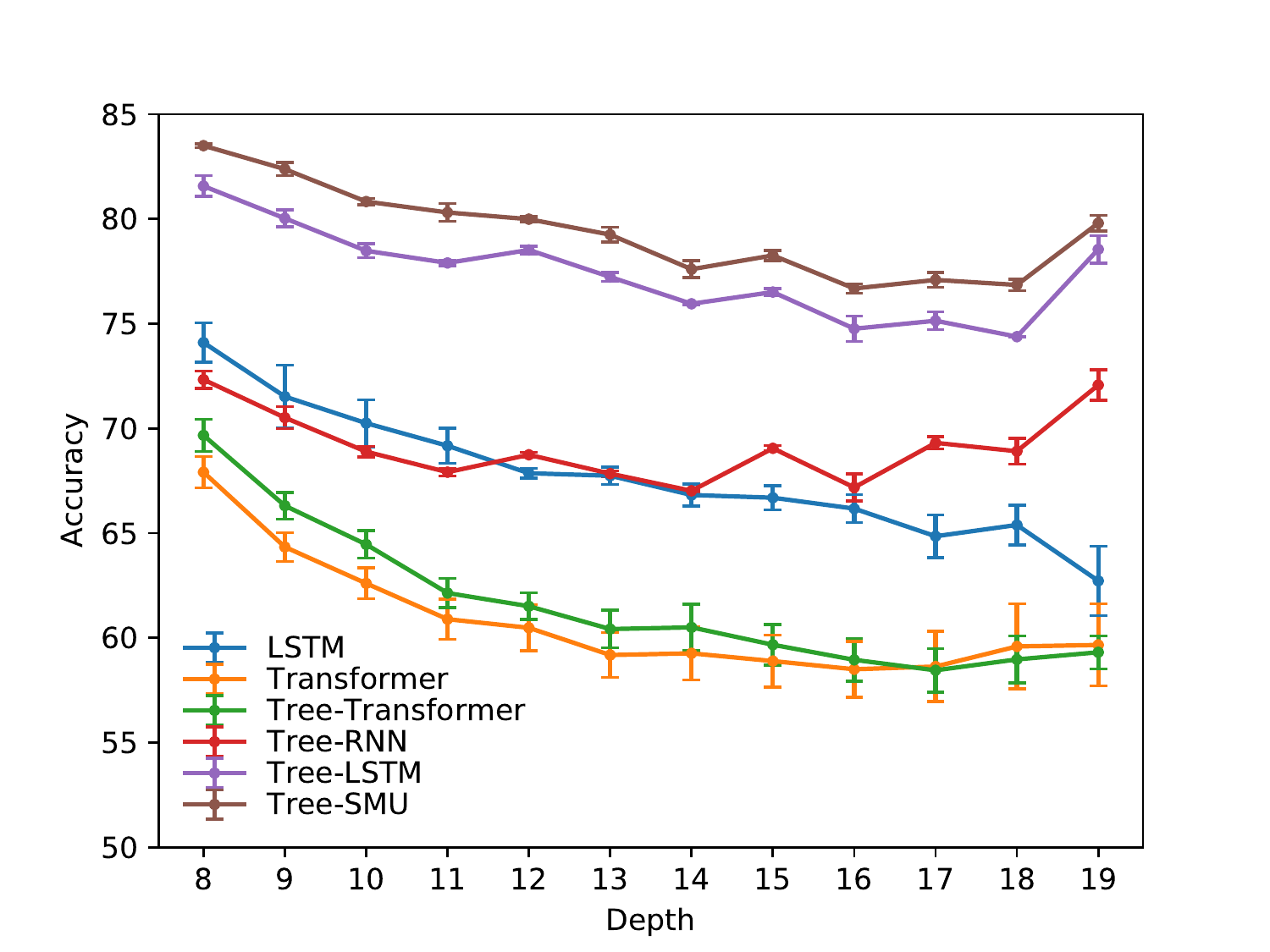}
    \caption{Equation Verification Experiment}
    \label{fig:depth_breakdown}
    \end{subfigure}
    \qquad
    \begin{subfigure}{0.4\textwidth}
    \centering
    \includegraphics[width=\textwidth]{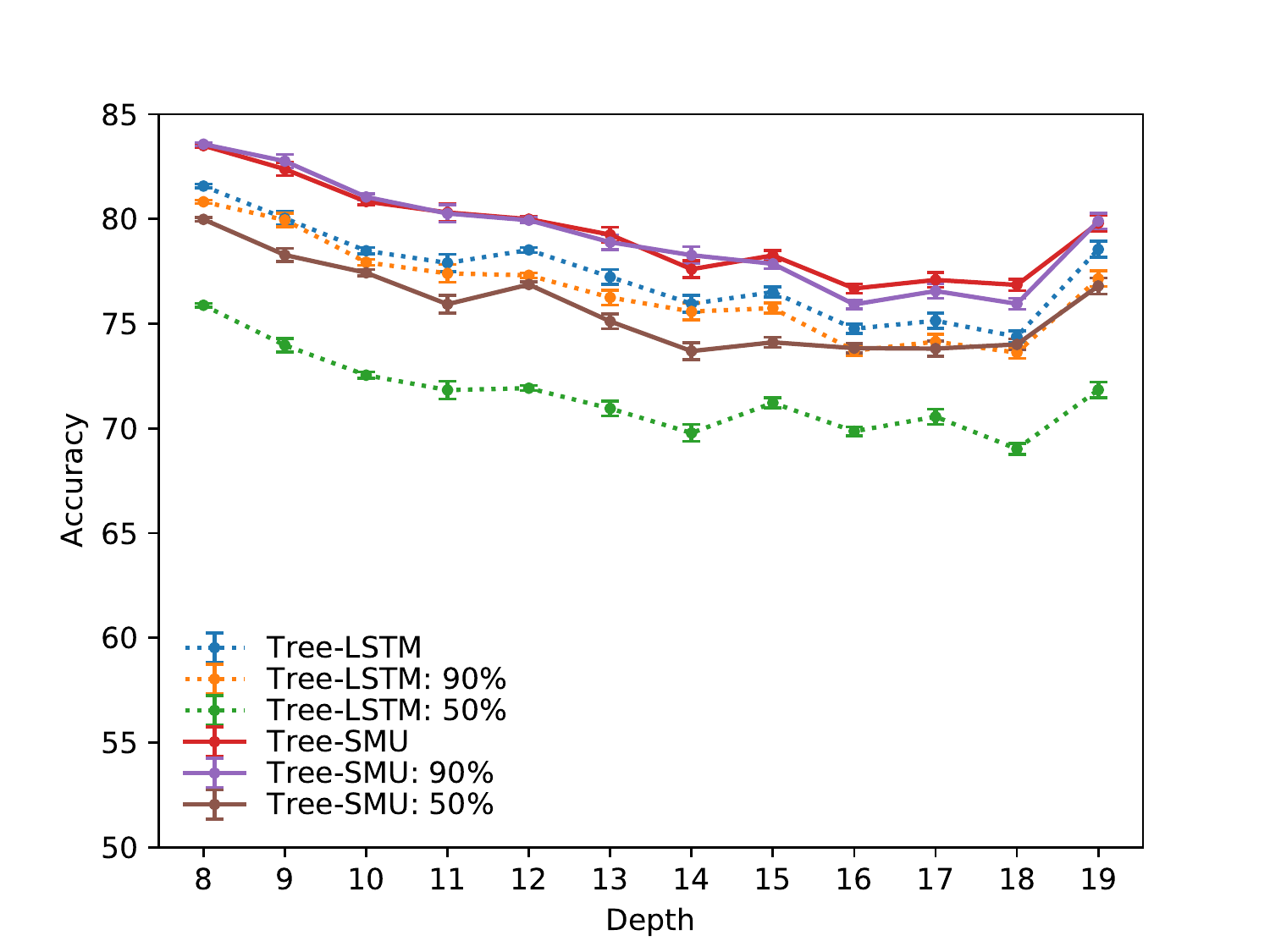}
    \caption{Sample efficiency}
    \label{fig:sample_eff_new}
    \end{subfigure}
    \caption{\textbf{Productivity Test} breakdown of model accuracy across different depths for the stack-LSTM models and baselines Tree-RNN and Tree-LSTM. The percentage in Fig.~\ref{fig:sample_eff_new} indicates the percentage of the training data used for training each model.}
    \vspace{-1.5em}
\end{figure}
\vspace{-1em}
\paragraph{Compositionality Tests}
In order to evaluate the compositional generalization of our proposed model, we use four of the five tests proposed by \citet{hupkes2020compositionality}. 
\begin{itemize}[leftmargin=*, topsep=0pt, itemsep=0ex ,partopsep=0ex]
    \item[] \textbf{Localism} states that the meaning of complex expression derives from the meanings of its constituents and how they are combined. To see if models are able to capture localism, we train the models on deep expressions (depth 5-13) and test them on shallow expressions (depth 1-4).
    \item[] \textbf{Productivity} has to do with the infinite nature of compositionality. For example, we can construct arbitrarily deep (potentially infinite) equations using a finite number of functions, symbols and numbers and ideally our model should generalizes to them. In order to test for productivity, we train our models on equations of depth 1-7 and test the model on equations of depth greater than 7.
    \item[] \textbf{Substitutivity} has to do with preserving the meaning of equivalent expressions such that replacing a sub-expression with its equivalent form does not change the meaning of the expression as a whole. In order to test for substitutivity, we look at t-SNE plots \citep{maaten2008visualizing} of the learned representations of sub-expressions and show that equivalent expressions (e.g. $y\times0$ and $(1\times x)\times 0$) map close to each other in the vector space. 
    \item[] \textbf{Systematicity} refers to generalizing to the recombination of known parts and rules to form other complex expressions. In order to assess this, 
    we test the model on test data equations with depth 1-7, and train it on training data equation with depth 1-7.
\end{itemize}
\paragraph{Baselines} We use several baselines listed below to validate our experiments. 
All the recursive models (including Tree-\SMU) perform a binary classification by optimizing the cross-entropy loss at the output which is the root (equality node). At the root, the models compute the 
dot product of the output embeddings of the right and the left sub-tree. The input of the recursive networks are the terminal nodes (leaves) of the equations that consist of symbols, representing variables in the equation, and numbers. The leaves of the recursive networks are embedding layers 
that embed the symbols and numbers in the equation. 
The parameters of the nodes that have the same functionality are shared. For example, all the addition functions use the same set of parameters.
Recurrent models input the data sequentially with parentheses.
\vspace{-0.5em}
\begin{itemize}[leftmargin=*,topsep=0pt,itemsep=-0.2ex]
    \item[] {\bf{Majority Class}} is a classification approach that always predicts the majority class. 
    \item[] {\bf{LSTM}} is a Long Short Term Memory network \cite{hochreiter1997long}.
    \item[] {\bf{Transformer}} is the transformer architecture of \citet{vaswani2017attention}
    \item[] {\bf{Tree Transformer}} a transformer with tree-structured positional encodings \citep{shiv2019novel}
    \item[] {\bf{Tree-RNN}} is a recursive neural network whose nodes are 2-layer feed-forward networks. 
    \item[] {\bf{Tree-LSTM}} is the Tree-LSTM network \cite{tai2015improved} presented in Section \ref{sec:background}.
\end{itemize}
\begin{figure}
    \centering
    \begin{subfigure}{0.4\textwidth}
    \centering
    \includegraphics[width=\textwidth]{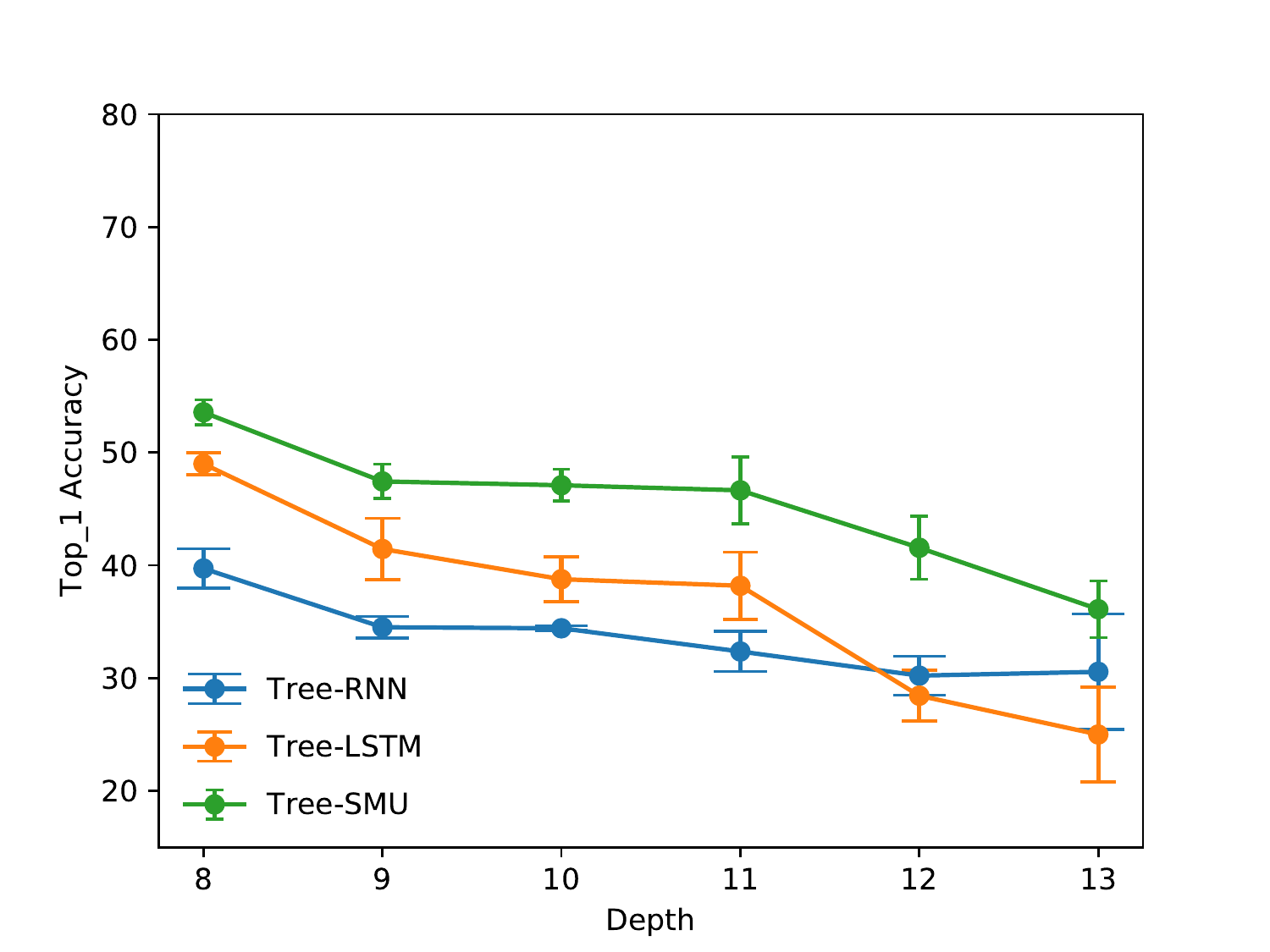}
    \caption{Top 1 Accuracy }
    \label{fig:top1}
    \end{subfigure}
    \qquad
    \begin{subfigure}{0.4\textwidth}
    \centering
    \includegraphics[width=\textwidth]{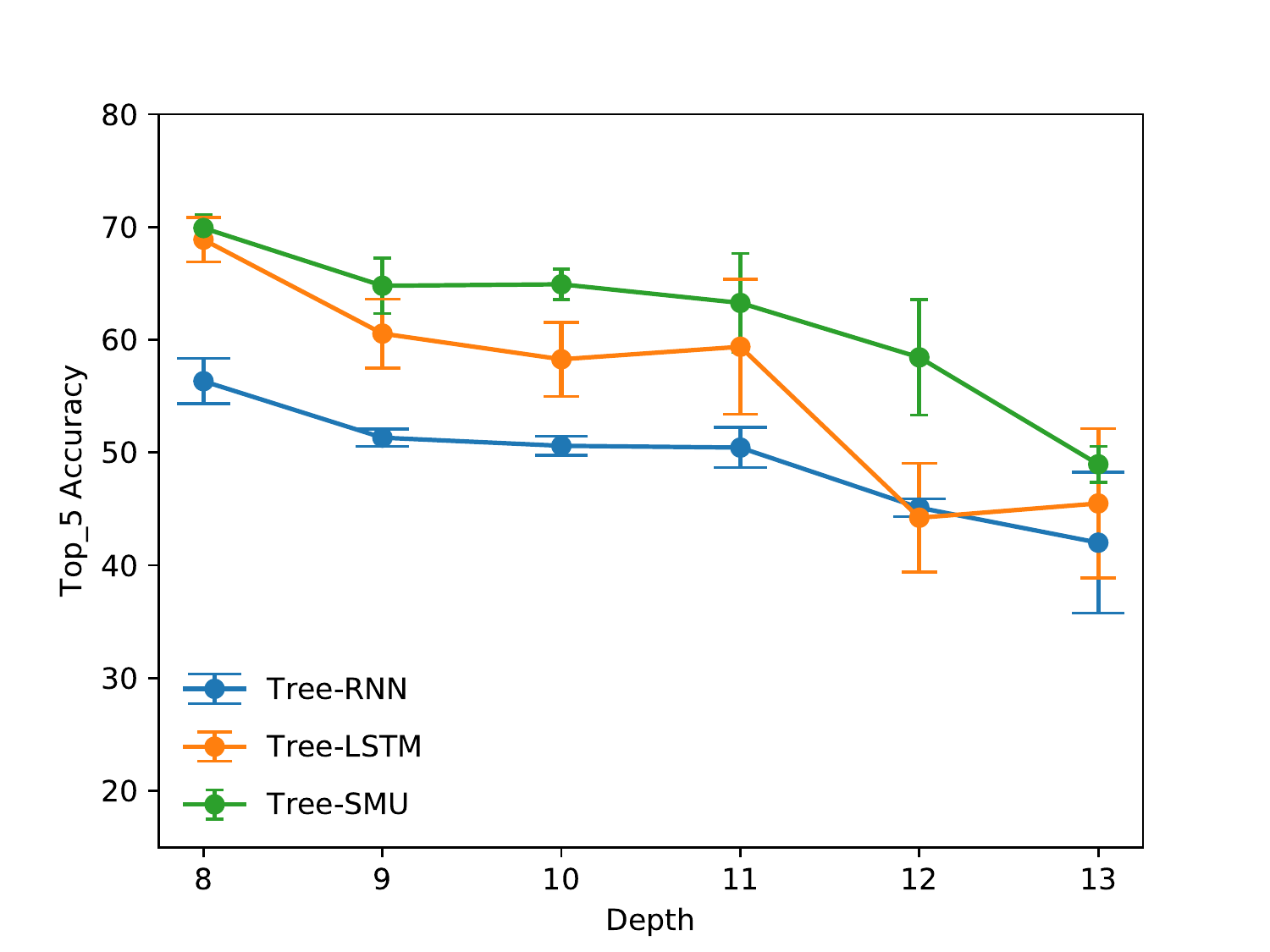}
    \caption{Top 5 Accuracy}
    \label{fig:top5}
    \end{subfigure}
    \caption{\textbf{Productivity Test} for Equation Completion. Top-$K$ accuracy metrics breakdown in terms of the test data depth.}
    \label{fig:topk}
    \vspace{-2em}
\end{figure}


{\bf{Evaluation metrics:}} For equation verification, we use the binary classification accuracy metric.
For equation verification, we use the top-$K$ accuracy \citep{arabshahi2018combining} which is the percentage of samples for which there is at least one correct match for the blank in the top-$K$ predictions.
\paragraph{Implementation Details}
\label{sec:implementation}
The models are implemented in PyTorch \citep{paszke2017automatic}. 
All the tree models use the Adam optimizer \citep{kingma2014adam} with $\beta_1=0.9$, $\beta_2=0.999$, learning rate $0.001$, and weight decay $0.00001$ with batch size $32$. 
We do a grid search for all the models to find the optimal hidden dimension and dropout rate in the range $[50,55,60,80,100,120]$ and $[0,0.1,0.15,0.2,0.25]$, respectively. Tree-\SMU's stack size is tuned in the range $[1,2,3,4,5,7,14]$ and the best accuracy is obtained for stack size $2$. 
All the models are run for three different seeds. We report the average and sample standard deviation of these three seeds. We choose the models based on the best accuracy on the validation data (containing equations of depths 1-7)\footnote{The code and data for re-producing all the experiments will be released upon acceptance. We have implemented a dynamic batching algorithm that achieves a runtime comparable with Transformers}. 

\vspace{-1em}
\section{Results and Discussion}
\label{sec:results}
\vspace{-1em}
\subsection{Compositionality Test Results}
\textbf{Localism} As shown in 
Tab.~\ref{tab:localism}, 
Tree-\SMU~significantly outperforms both Tree-RNN and Tree-LSTM achieving near perfect accuracy on the equation verification test set. This indicates that the stack memory is able to effectively capture and preserve locality, as claimed in the introduction.

\textbf{Productivity}
The productivity test, evaluates the stack's capability of capturing global long-range dependencies. In this test, the model is evaluated on zero-shot generalization to deeper mathematical expressions. This is a harder task compared to localism. We run the productivity test on both the equation completion and the equation verification tasks. 

\emph{Equation Verification:} 
Table \ref{tab:overall_results} shows the overall accuracy of all the baselines on equations of depth 8-19. The break-down of the accuracy vs. the equation depth is shown in Fig.~\ref{fig:depth_breakdown}. As it can be seen, Tree-\SMU~consistently outperforms all the baselines on all depths. This indicates that stack effectively captures global long-range dependencies, allowing the model to compositionally generalize to deeper and harder mathematical expressions. However, this task is harder and the improvement margin of this test is smaller compared to the localism test. 

\emph{Equation Completion:} In order to perform equation completion, we use the model trained on equation verification to predict blank fillers for an input equation that has a blank. The test data, different from the equation verification test data, is generated by randomly substituting sub-trees of depth 1 or 2 in equations of depth 8-13 with a blank leaf. In order to make predictions for the blank, we generate candidates of depth 1 and 2 from the data vocabulary and use the trained models to rank the candidates. Figures~\ref{fig:top1} and \ref{fig:top5} show the top-1 and top-5 accuracy for the recursive models. As it can be seen, the performance of Tree-\SMU~is consistently better on all depths compared to Tree-LSTM and Tree-RNN. The performance of the recurrent models were poor and are not shown in these plots.

\begin{figure}[t]
    \centering
    \begin{subfigure}{0.35\textwidth}
    \includegraphics[clip, trim=3.5cm 2cm 3cm 2cm, width=\textwidth]{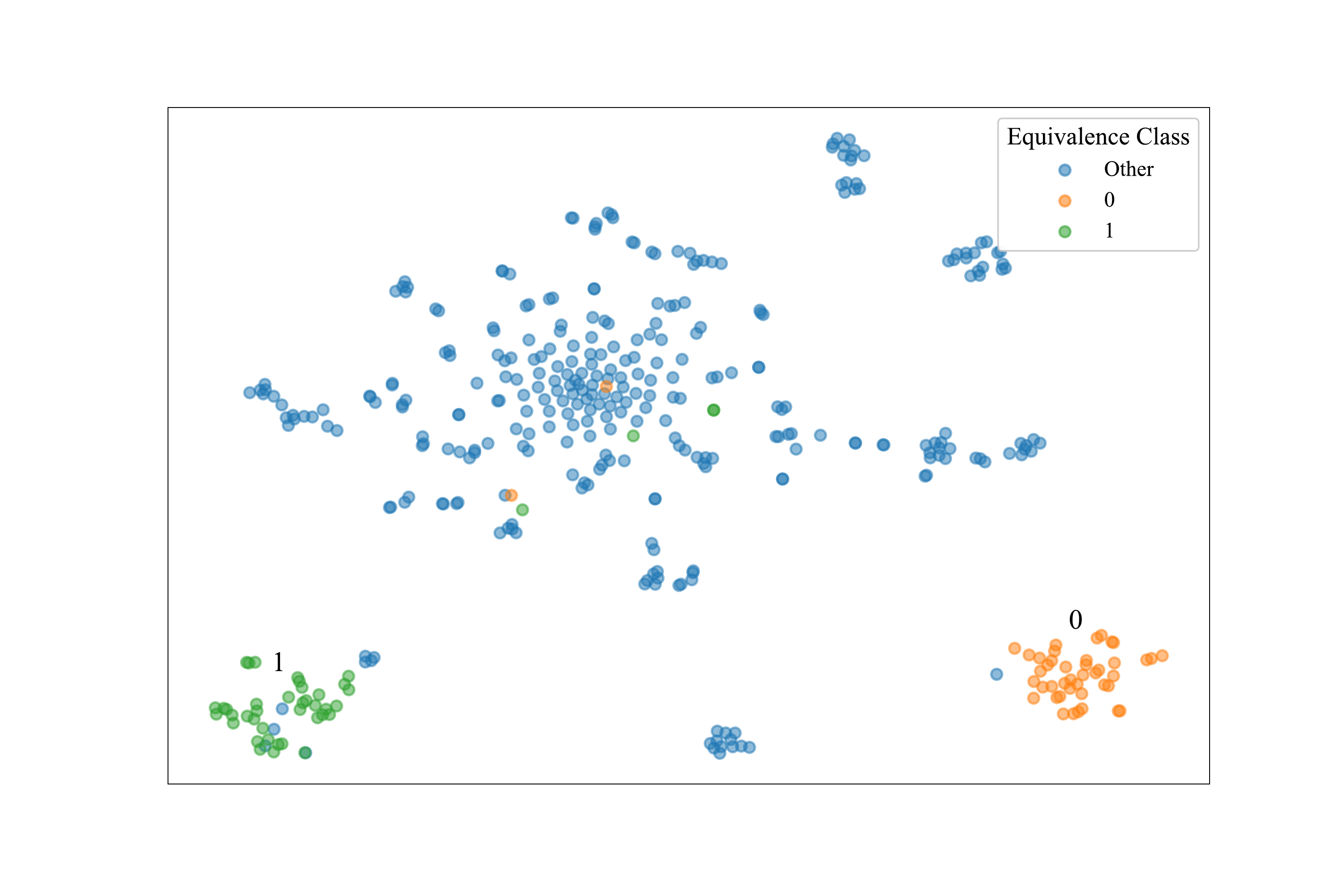}
    \caption{Tree-\SMU~learned representations}
    \label{fig:tsne-tsmu}
    \end{subfigure}
    \qquad\qquad
    \begin{subfigure}{0.35\textwidth}
    \includegraphics[clip, trim=3.5cm 2cm 3cm 2cm, width=\textwidth]{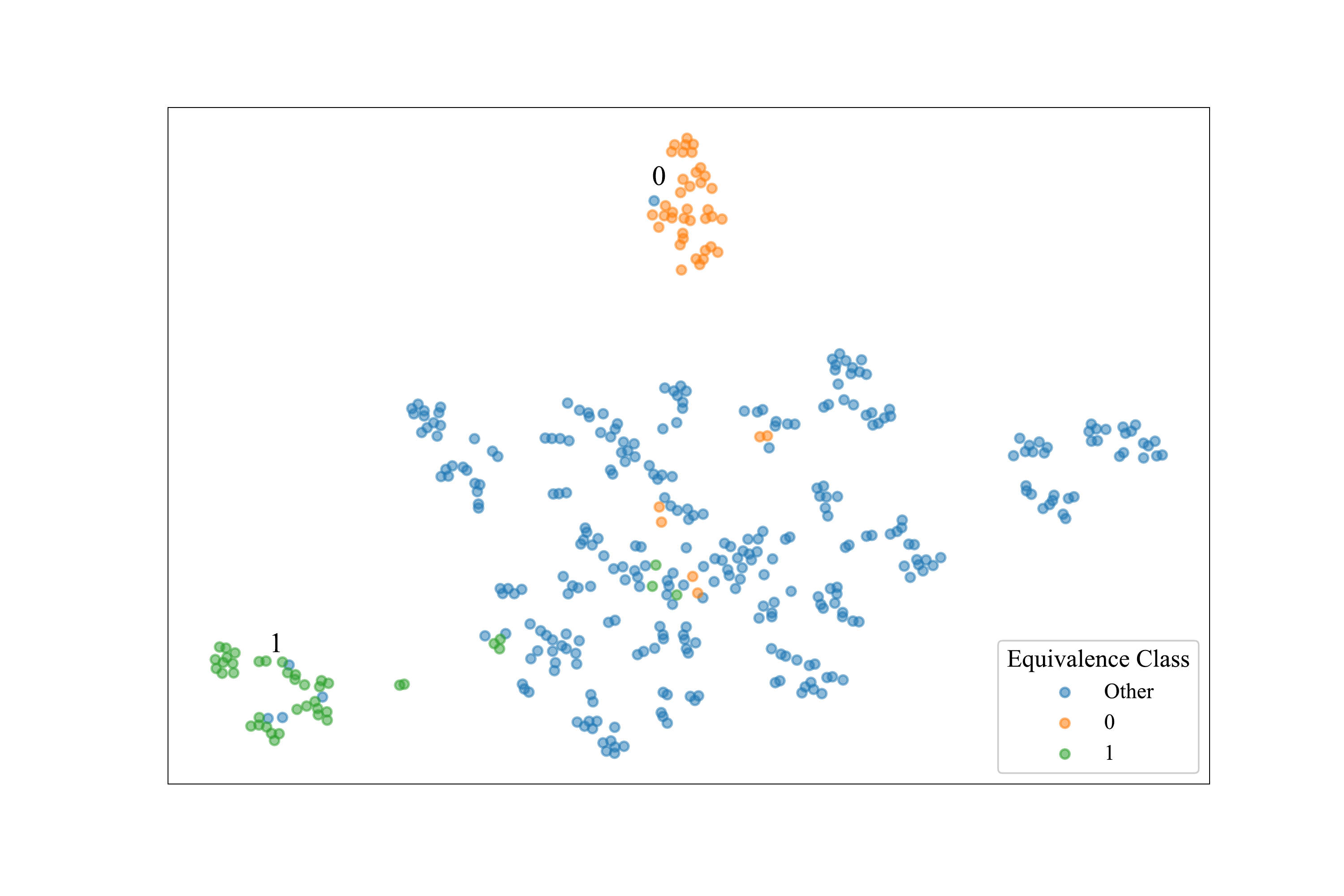}
    \caption{Tree-LSTM learned representations}
    \label{fig:tsne-trnn}
    \end{subfigure}
    \caption{\textbf{Substitutivity Test}: Learned representations of Tree-\SMU~and Tree-LSTM colored by expressions' class label. Both the tree models are able to cluster equivalent expressions. However, Tree-\SMU~is able to learn a much smoother representation per class label, whereas Tree-LSTM is more sensitive to irrelevant syntactic details. For example, the two sub-clusters of class $1$ in Fig \ref{fig:tsne-trnn} group expressions raised to the power of $0$ in the top sub-cluster and 1 raised to the power of an expression in the bottom sub-cluster.} 
    \label{fig:tsne}
    \vspace{-1.5em}
\end{figure}

\textbf{Substitutivity} The t-SNE plots of the representations learned by Tree-LSTM and Tree-\SMU~are shown in Fig.~\ref{fig:tsne}. Each point on the t-SNE plot is a mathematical expression with varying depths ranging from 1-3. The hyperparameters for both t-SNE plots are the same. We have highlighted two of the clustered expressions with their equivalence class label, viz., $0$ and $1$. These include expressions that evaluate to these numbers (e.g., $1\times (x \times 0)$ is in the $0$ cluster). 

As it can be seen in the plots, both models are able to form clusters of equivalence classes although they are not directly minimizing these class losses. However, Tree-\SMU~learns a much smoother representation per equivalence class. Moreover, Tree-LSTM forms sub-clusters that are sensitive to irrelevant syntactic details. For example, the top sub-cluster of $1$ in Fig \ref{fig:tsne-trnn} is the group of expressions raised to the power of $0$ (e.g., $x^{0}$) and the bottom sub-cluster is the group of $1$ raised to the power of an expression (e.g., $1^{1\times y}$). Tree-RNN is even more sensitive to these syntactic details (Fig.~\ref{fig:tsne-trnn_real}, Appendix). For example, Tree-RNN learns distinct sub-clusters for equivalence class $0$ that are grouped by expressions multiplied by $0$ from the right (e.g., $\frac{1}{2}\times0$), and expression multiplied by $0$ from the left (e.g., $0 \times -1$).
Another example of irrelevant syntactic detail captured by Tree-RNN is sensitivity to sub-expression depth. Therefore,  if we substitute a sub-expression with a semantically equivalent one, it is likely for the meaning of the expression as a whole to change due to Tree-RNN's sensitivity to irrelevant syntactic details. Whereas Tree-\SMU~passes the substitutivity test.

\textbf{Systematicity} The systematicity test results are shown in Tab.~\ref{tab:overall_results_full}, column 3. As it can be seen the accuracy of Tree-LSTM is comparable with Tree-\SMU. This compositional generalization test is easier compared with the previous ones because the training and test datasets are more similar in this case. Therefore, Tree-\SMU's is more robust to changes in the compositionality of the test set compared to Tree-LSTM. This also indicates that Tree-LSTM is more likely to learn irrelevant artifacts in the data that hurts its performance more under data distribution shifts.


\vspace{-1em}
\subsection{Sample Efficiency}
\vspace{-1em}
Finally, Figure~\ref{fig:sample_eff_new} shows that Tree-\SMU~also has a better sample efficiency compared with Tree-LSTM. This figure shows the performance on the entire productivity test data for equation verification, but the models are trained on a sub-sample of the training data. The sub-sampling percentage is shown in the figure. As shown, the accuracy of Tree-\SMU~trained on 50\% of the data is comparable to (at some depth slightly lower than) the accuracy of Tree-LSTM trained on the full dataset. However, the performance of Tree-LSTM significantly degrades when trained on 50\% of the training data. Moreover, even when Tree-\SMU~is trained using 90\% of the data, it still outperforms Tree-LSTM.

\vspace{-1.5em}
\section{Related Work}
\vspace{-1.5em}
Recursive neural networks have been used to model compositional data in many applications e.g., natural scene classification \citep{socher2011parsing}, sentiment classification, Semantic Relatedness and syntactic parsing \citep{tai2015improved,socher2011parsing}, neural programming, and logic \citep{allamanis2017learning,zaremba2014learning, evans2018can}. In all these problems, there is an inherent compositional structure nested in the data. Recursive neural networks integrate this symbolic domain knowledge into their architecture and as a result achieve a significantly better generalization performance. 
However, we show that standard recursive neural networks do not generalize in a zero-shot manner to unseen compositions mainly because of error propagation in the network. Therefore, we propose a novel recursive neural network, \SMUlong~that shows strong zero-shot generalization compared with our baselines.



\begin{table}[t]
\begin{minipage}[t]{0.55\textwidth}
  \caption{\textbf{Productivity and Systematicity Tests} Equation Verification, Overall model Accuracy on train and test datasets }
\label{tab:overall_results}
\resizebox{\textwidth}{!}{%
\centering
\begin{tabular}{lcccc}
\toprule 
\textbf{Approach} & 
	\textbf{\thead[c]{Train\\(Depths 1-7)}} & 
	\textbf {\thead[c]{Validation\\(Depths 1-7)}} & 
	\textbf{\thead[c]{Systematicity Test\\(Depth 1-7)}} &
    \textbf {\thead[c]{Productivity Test\\(Depths 8-19)}} \\
\midrule
Majority Class &  58.12 & 56.67 & 60.40 & 51.71 \\
\addlinespace
LSTM & 85.62 & 
79.48$\pm{4.53}$
& $79.24 \pm{0.006}$ & 68.36$\pm{0.42}$ \\
Transformer & 81.26 & $74.24 \pm{0.02}$ & 76.45$\pm{0.42}$
& 61.05$\pm{1.53}$ \\
Tree Transformer & 84.08 & $74.90 \pm{0.02}$ & 77.80$\pm{0.46}$
& 62.12$\pm{1.06}$ \\

\addlinespace
Tree-RNN & $99.11$ & $89.45 \pm{0.08}$  & $72.41 \pm{0.01}$ & $68.95\pm{0.24}$ \\
Tree-LSTM & $99.86$ & $93.05\pm{0.12}$
 & $83.06 \pm{0.01}$ & $77.58\pm{0.19}$ \\
\addlinespace
Tree-SMU & 99.59 & $93.52\pm{0.28}$ & $\mathbf{83.29 \pm{0.01}}$ & $\mathbf{79.57\pm{0.16}}$ \\
\bottomrule
\end{tabular}
}
\end{minipage}
~~
\begin{minipage}[t]{0.40\textwidth}
\caption{\textbf{Localism Test} Equation Verification, Overall model Accuracy on train and test datasets}
\label{tab:localism}
\resizebox{\textwidth}{!}{%
\centering
\begin{tabular}{lccc}
\toprule 
\textbf{Approach} & 
	\textbf{\thead[c]{Train\\(Depths 5-13)}} & 
	\textbf {\thead[c]{Validation\\(Depths 5-13)}} & 
    \textbf {\thead[c]{Test\\(Depths 1-4)}} \\
\midrule
Majority Class &  $55.08$ & $56.45$ & $60.88$\\
\addlinespace
Tree-RNN & $99.07$ & $88.33 \pm{0.37}$  & $85.36\pm{0.35}$ \\
Tree-LSTM & $99.79$ & $91.70\pm{0.0.28}$
 & $91.58\pm{0.11}$ \\
\addlinespace
Tree-SMU & $95.04$ & $92.78\pm{0.09}$ & $\mathbf{98.86\pm{0.07}}$ \\
\bottomrule
\end{tabular}
}
\end{minipage}
\vspace{-1em}
\end{table}

\SMUlong~is a recursive network with an extended structured memory. Recently, there have been attempts to provide a global memory to recurrent neural models that plays the role of a working memory and can be used to store information to and read information from \citep{graves2014neural,weston2015, grefenstette2015learning, joulin2015inferring}. Memory networks and their differentiable counterpart \citep{weston2015, sukhbaatar2015end} store instances of the input data into an external memory that can later be read through their recurrent neural network architecture. Neural Programmer Interpreters augment their underlying recurrent LSTM core with a key-value pair style memory and they additionaly enable read and write operations for accessing it \citep{reed2015neural,cai2017making}. Neural Turing Machines \citep{graves2014neural} define soft read and write operations so that a recurrent controller unit can access this memory for read and write operations. Another line of research proposes to augment recurrent neural networks with specific data structures such as stacks and queues \citep{das1992learning, dyer2015transition, sun2017neural, joulin2015inferring, grefenstette2015learning, mali2019neural}. These works provide an external memory for the neural network to access, whereas our proposed model integrates the memory within the cell architecture and does not treat the memory as an external element. Our design experiments showed that recursive neural networks do not learn to use the memory if it is an external component. Therefore, a trivial extension of works such as \citet{joulin2015inferring} to tree-structured neural networks does not work in practice. 

Despite the amount of effort spent on augmenting recurrent neural networks, to the best of our knowledge, there has been no attempt to increase the memory capacity of recursive networks, which will allow them to extrapolate to harder problem instances.
Therefore, inspired by the recent attempts to augment recurrent neural networks with stacks, we propose Tree Stack Memory Units, a recursive neural network that consists of differentiable stacks. We propose novel soft push and pop operations to fill the memory of each Stack Memory Unit using the stacks and states of its children. 
It is worth noting that a trivial extension of stack augmented recurrent neural networks such as \cite{joulin2015inferring} results in the stack-augmented Tree-RNN structure presented in the Appendix. We show in our experiments that this trivial extension does not work very well. 

In a parallel research direction, an episodic memory was presented for question answering applications \cite{kumar2016ask}. This is different from the symbolic way of defining memory. Another different line of work are graph memory networks and tree memory networks \cite{pham2018graph, fernando2018tree} which construct a memory with a specific structure. These works are different from our proposed recursive neural network which has an increased memory capacity due to an increase in the memory capacity of each cell in the recursive architecture. 

\vspace{-1em}
\section{Conclusions}
\vspace{-1em}
In this paper, we study the problem of zero-shot generalization of neural networks to novel compositions of domain concepts. This problem is referred to as compositional generalization and it currently a challenge for state-of-the-art neural networks such as transformers and Tree-LSTMs. In this paper, we propose \SMUlong~(Tree-\SMU) to enable compositional generalization. Tree-\SMU~is a novel recursive neural network with an extended memory capacity compared to other recursive neural networks such as Tree-LSTM. The stack memory provides an error correction mechanism thriygh gives each node indirect access to its descendants acting as an error correction mechanism. 
Each node in Tree-\SMU~has a built in differentiable stack memory. \SMU~learns to read from and write to its memory using its soft push and pop gates. We show that Tree-\SMU~achieves strong compositional generalization compared to baselines such as transformers, tree transformers and Tree-LSTMs for mathematical reasoning.

\clearpage
\section*{Appendix}

\subsection*{Data distribution}

\begin{figure}[h]
    \centering
    \includegraphics[width=1\textwidth]{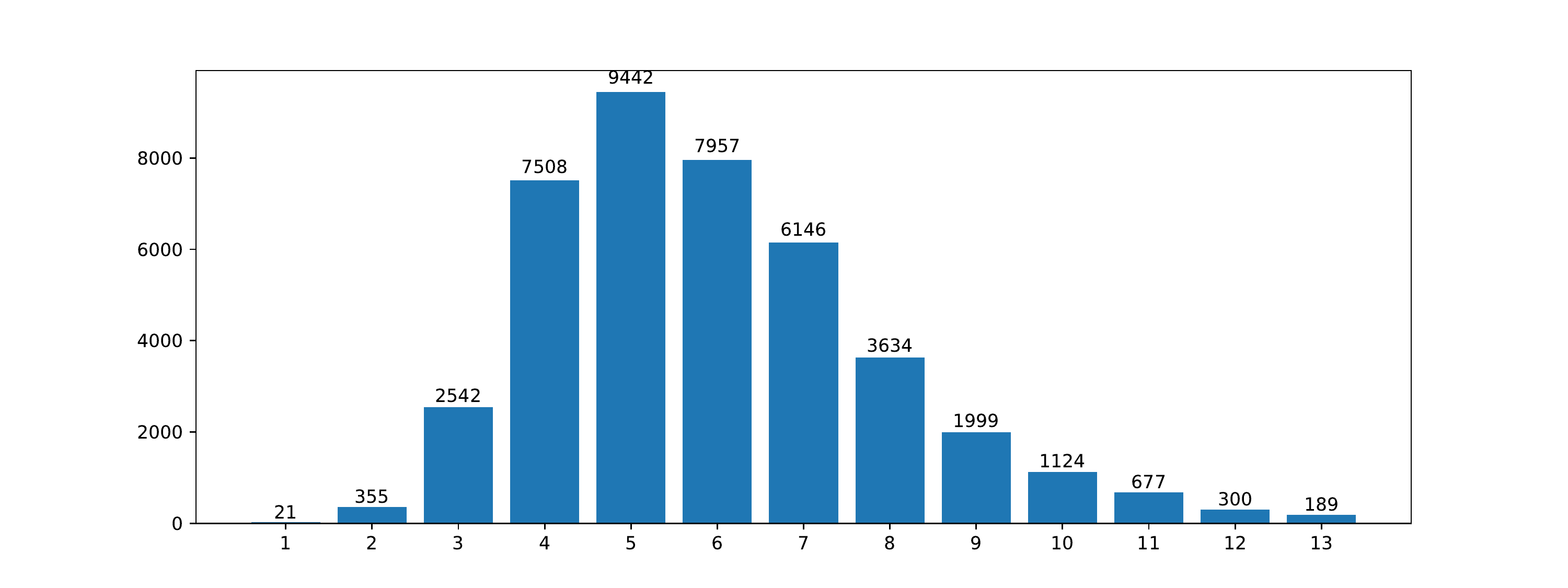}
    \caption{Number of equations in the train/validation 40k data broken down by their depth.}
    \label{fig:data_stats}
    \vspace{-1em}
\end{figure}

A data generation strategy for these tasks was presented in \cite{arabshahi2018combining} and we use that to generate symbolic mathematical equations of up to depth 13. 
We generate $41,894$ equations of different depths (Figure~\ref{fig:data_stats} gives the number of equations in each depth).
This dataset is approximately balanced with a total of $56\%$ correct and $44\%$ incorrect equations. This dataset is used for training and validation in the equation verification experiments. The test data is generated with a different seed and different hyper-parameters. This was done to encourage the data generation to generate a depth balanced dataset. The test dataset has about 200k equation of depths 1-19 with roughly $10,000$ equations per depth (excluding depths 1 and 2). Equation verification models are run only once on the large test data and reported in the paper. Example equations from the training dataset are shown in Table \ref{tab:examples}.


\begin{table*}[h]
  \caption{\textbf{Examples of generated equations in the dataset}}
\label{tab:examples}
\centering
\resizebox{1\textwidth}{!}{%
\begin{tabular}{llc}
\toprule
\bf Example & \bf Label & \bf Depth \\ 
\midrule
\addlinespace
$(\sqrt{1} \times 1 \times y) + x = (1 \times y) + x$ & Correct & 4 \\
\addlinespace
$\sec(x + \pi) = (-1 \times \sec(\sec(x)))$ & Incorrect & 4 \\
\addlinespace
$y \times \Big(1^1 \times (3 + (-1 \times 4^{0 \times 1})) + x^1\Big) = y \times 2^0 \times (2 + x)$ & Correct & 8 \\
\addlinespace
$\sqrt{1 + (-1 \times (\cos(y + x))^{\sqrt{\csc(2)}} )} \times (\cos(y + x))^{-1} = \tan(y^1 + x)$ & Incorrect & 8\\
\addlinespace
$2^{-1} + \Big(-\frac{1}{2} \times \big (-1 \times \sqrt{1 + (-1 \times \sin^2(\sqrt{4} \times (\pi + (x \times -1))))} \big) \Big) + \cos^{\sqrt{4}}(x) = 1$ & Correct & 13 \\
\addlinespace
 $\big(\cos(y^1 + x) + z\big)^w = \big(\cos(x) \times \cos(0 + y) + \big(-1 \times \sqrt{1 + -1 \times \cos^{2}(y + 2\pi)}\; \big) \times \sin(x) + z \big)^w$  & Correct & 13  \\
 \addlinespace
 $\sin\Big( \sqrt{4}\;^{-1} \pi + (-1 \times \sec \big( \csc^2(x)^{-1} + \sin^2(1 + (-1 \times 1) + x + 2^{-1} \pi ) \big) \times x) \Big) = \cos(0 + x)$ & Incorrect & 13 \\
\addlinespace
\bottomrule
\end{tabular}}
\end{table*}

\subsection*{Extended Results and Discussion}

\paragraph{Productivity test} Table \ref{tab:overall_results_full} is the extended version of Table \ref{tab:overall_results} with additional evaluation metrics. These additional metrics are precision (Prec) and recall (Rcl) for the binary classification problem of equation verification. The results are shown along with the accuracy metrics reported in Table \ref{tab:overall_results}.

\begin{table}[h]
  \caption{\textbf{Productivity Test for Equation Verification:} Overall accuracy (Acc), precision (Prec) and recall (rcl) of the models on train and test datasets}
\label{tab:overall_results_full}
\resizebox{1\textwidth}{!}{%
\centering
\begin{tabular}{lccccccccc}
\toprule 
\multirow{2}{*}{\bf Approach} & 
	\multicolumn{3}{c}{\bf Train (Depths 1-7)} & 
	\multicolumn{3}{c}{\bf Validation (Depths 1-7)} & 
    \multicolumn{3}{c}{\bf Test (Depths 8-19)} \\
\cmidrule(lr){2-4}
\cmidrule(lr){5-7}
\cmidrule(lr){8-10}
 & \bf Acc & \bf Prec & \bf Rcl & \bf Acc & \bf Prec & \bf Rcl & \bf Acc & \bf Prec & \bf Rcl \\
\midrule
Majority Class &  58.12 & - & - & 56.67 & - & -
& 51.71 & - & - \\
\addlinespace
LSTM & 85.62 & 82.14 & 97.15 & 
79.48$\pm{4.53}$ & 76.18$\pm{5.01}$ & 93.71$\pm{0.80}$
& 68.36$\pm{0.42}$ & 72.87$\pm{7.01}$ & 60.25$\pm{14.89}$ \\
Transformer & 81.26 & 78.36 & 93.67 & 76.45$\pm{0.42}$ & 73.63$\pm{0.80}$ & 91.11$\pm{1.93}$
& 61.05$\pm{1.53}$ & 59.96$\pm{1.50}$ & 60.35$\pm{7.28}$ \\
Tree Transformer & 84.08 & 80.93 & 95.03 & 77.80$\pm{0.46}$ & 74.97$\pm{0.87}$ & 91.36$\pm{1.25}$
& 62.12$\pm{1.06}$ & 60.63$\pm{1.54}$ & 63.68$\pm{2.93}$ \\

\addlinespace
Tree-RNN & $99.11$ & $98.92$ & $99.56$ & $89.45 \pm{0.08}$  & $88.47\pm{0.27}$ & $93.57\pm{0.40}$
& $68.95\pm{0.24}$ & $\mathbf{81.82\pm{1.42}}$ & $46.60\pm{1.74}$ \\
Tree-LSTM & $99.86$ & $99.80$ & $99.96$ & $93.05\pm{0.12}$ & $89.67\pm{1.32}$ & $98.36\pm{0.51}$
 & $77.58\pm{0.19}$ & $79.50\pm{0.22}$ & $72.69\pm{0.25}$ \\
\addlinespace
Tree-SMU & 99.59 & 99.31 & 99.98 & $93.52\pm{0.28}$ & $91.08\pm{0.49}$ &
$98.19\pm{0.10}$ & $\mathbf{79.57\pm{0.16}}$ & $80.56\pm{2.11}$ & $\mathbf{76.63\pm{3.15}}$ \\
\bottomrule
\end{tabular}
}
\end{table}

\paragraph{Stack size ablation} 
In this experiment, we train Tree-\SMU~for various stack sizes $p=[1,2,3,7,14]$. The model is trained on equations of depth 1-7 from the training set and evaluated on equations of depth 1-13 from the validation set. The purpose of this experiment is to indicate how many of the stack rows are being used by the model for equations of different depth. The results are shown in Table \ref{tab:stack_size_ablation}. We say that a stack row is being used if the P2 norm of that row has a value higher than a threshold $\tau = 0.001$. As shown in the Table, as the equations' depth increase, more of the stack rows are being used. This might indicate that the \SMU~cells are using the extended memory capacity to capture long-range dependencies and overcome error-propagation.

\begin{table}[]
    \centering
    \caption{Average stack usage for different stack sizes.}
    \label{tab:stack_size_ablation}
    \begin{tabular}{cc|ccccccccccc}
    \toprule
        \multicolumn{2}{c|}{\multirow{2}{*}{Stack Size}} & \multicolumn{11}{c}{Average Stack Usage broken down by Data Depth}\\
        \multicolumn{2}{c|}{}& 3 & 4 & 5 & 6 & 7 & 8 & 9 & 10 & 11 & 12 & 13  \\
        \midrule
        \multicolumn{2}{c|}{1}  &$0.05$ & $0.19$ & $0.41$ &$0.63$ &$0.80$ &$0.89$ &$0.89$&$0.91$ &$0.97$ &$0.89$ &$0.82$  \\
        \multicolumn{2}{c|}{2}  &$0.05$ &$0.31$ &$0.74$ &$1.16$ &$1.45$ &$1.63$ &$1.69$ &$1.66$ &$1.77$ &$1.73$ &$1.54$  \\
        \multicolumn{2}{c|}{3}  &$0.05$ &$0.31$ &$0.93$ &$1.55$ &$1.97$ &$2.18$ &$2.32$ &$2.23$ &$2.35$ &$2.39$ &$2.13$  \\
       \multicolumn{2}{c|}{7}  &$0.05$ &$0.31$ &$0.93$ &$1.73$ &$2.47$ &$2.82$ &$3.21$ &$3.00$ &$3.14$ &$3.51$ &$3.31$ \\
       \multicolumn{2}{c|}{14} & $0.05$ &$0.31$ &$0.93$ &$1.73$ &$2.47$ &$2.82$ &$3.21$ &$3.05$ &$3.23$ &$3.76$ &$3.53$ \\
       \bottomrule
    \end{tabular}
\end{table}

\begin{figure}[h]
    \centering
    \includegraphics[clip, trim=3.5cm 2cm 3cm 2cm, width=0.6\textwidth]{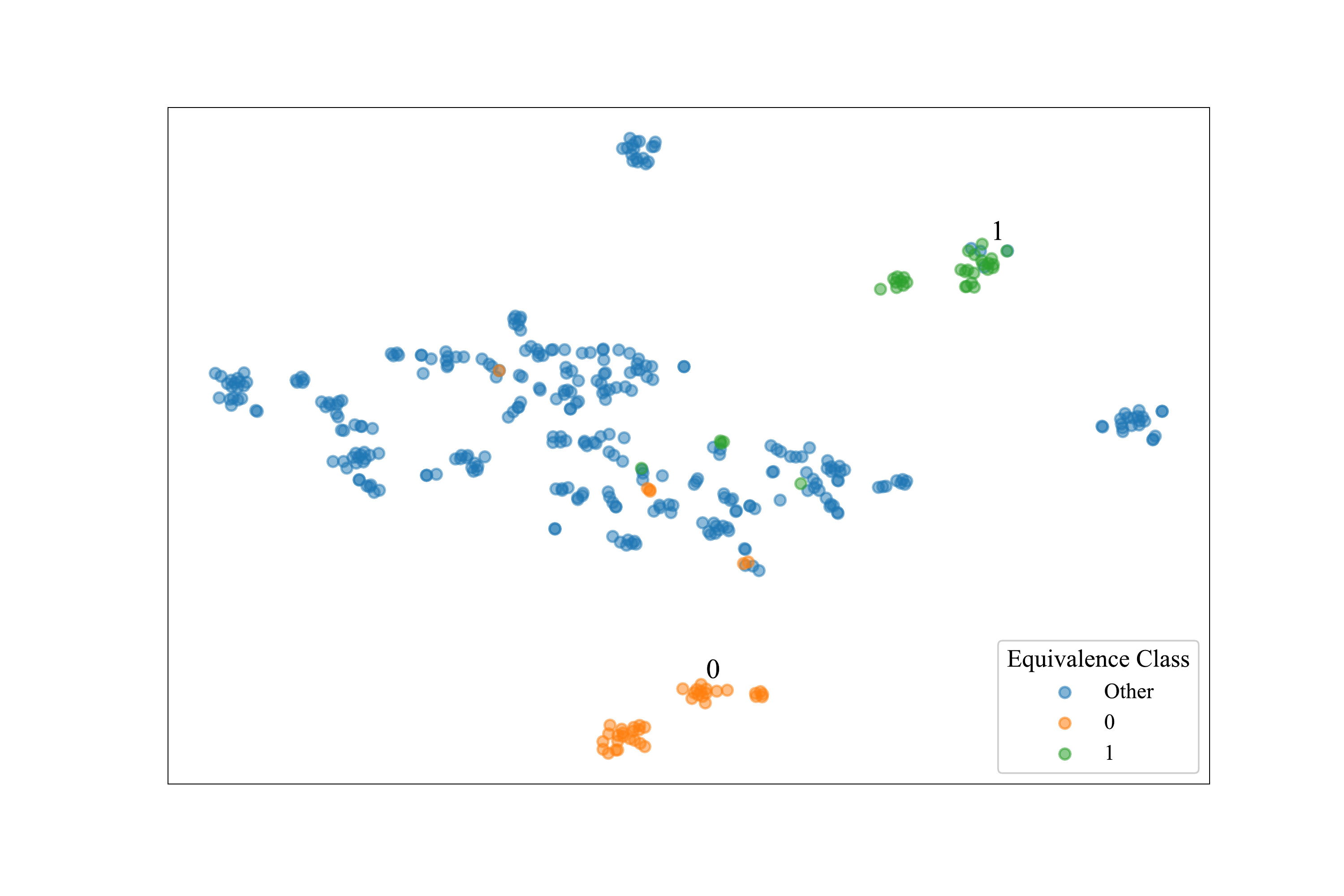}
    \caption{\textbf{Substitutivity Test}: Learned representations of Tree-RNN. The model is able to cluster equivalent expressions. However, Tree-RNN is very sensitive to irrelevant syntactic details. For example, the two sub-clusters of class $0$ group expressions multiplied by 0 on the left (left sub-cluster) vs. on the right (right sub-cluster).} %
    \label{fig:tsne-trnn_real}
    \vspace{-1.5em}
\end{figure}

\paragraph{T-SNE plots} Finally, we show the t-SNE plot visualizations for the tree-RNN in Figure \ref{fig:tsne-trnn_real}. The learned representations by Tree-RNN are even more sensitive to irrelevant syntactic details compared with Tree-LSTM and Tree-\SMU. These variations in the learned representation of equivalent sub-expressions, results in errors that propagate as expressions get deeper.

\end{document}